# Unraveling Cold Start Enigmas in Predictive Analytics for OTT Media: Synergistic Meta-Insights and Multimodal Ensemble Mastery

## Kaushik Ganguly[1], Amit Patra[2]

[1]*AI/ML Architect, Cognizant Technology Solutions India Pvt. Ltd.*
[2]*Data Scientist, Cognizant Technology Solutions India Pvt. Ltd.*

------------------------------------------------------------------------***------------------------------------------------------------------------

**Abstract -** *The cold start problem is a common challenge in various domains, including media use cases such as predicting viewership for newly launched shows on Over-The-Top (OTT) platforms. In this study, we propose a generic approach to tackle cold start problems by leveraging metadata and employing multi-model ensemble techniques. Our methodology includes feature engineering, model selection, and an ensemble approach based on a weighted average of predictions. The performance of our proposed method is evaluated using various performance metrics. Our results indicate that the multi-model ensemble approach significantly improves prediction accuracy compared to individual models.*

## 1. Introduction

The cold start problem is a prevalent issue in predictive analytics, where there is limited or no historical data available for newly introduced items. This challenge is particularly evident in the media domain when predicting video views for newly launched shows on OTT platforms. Traditional machine learning algorithms often struggle to provide accurate predictions in such scenarios due to a lack of historical data. In this study, we aim to address the cold start problem by leveraging metadata and adopting a multi-model ensemble approach. We believe that this generic approach can be applied to various domains beyond media use cases.

## 2. Business Objective

The business objective of predicting video views for a web series in the OTT (Over-the-Top) domain is to optimize content production, marketing, and revenue generation strategies. Below are some possible ways in which video view predictions can help achieve these objectives:

- **Content Production Optimization:** Predicting video views can help OTT platforms decide which types of content to invest in and create. By analysing viewer behaviour and preferences, OTT platforms can use predictive analytics to identify the types of content that are most likely to be popular among their audience. This can help them make informed decisions about which shows to produce, what genres to focus on, and what kind of storylines to develop.

- **Marketing Optimization:** Predictive analytics can also help OTT platforms optimize their marketing efforts. By analysing user behaviour data, OTT platforms can identify the audience segments that are most likely to be interested in a particular show and use targeted marketing campaigns to reach these users. This can help increase awareness of the show and attract more viewers.

- **Revenue Generation Optimization:** Predicting video views can also help OTT platforms optimize their revenue generation strategies. By accurately predicting the number of views a show is likely to receive, OTT platforms can estimate how much revenue they are likely to generate from advertising, subscriptions, or other revenue streams. This can help them make informed decisions about how much to invest in content creation and marketing, and how to price their services.

Overall, predicting video views for web series in the OTT domain can help platforms optimize their content creation, marketing, and revenue generation strategies to achieve their business objectives.

## 3. Literature Review

Cold start problems have been widely explored in various domains, such as:
- collaborative filtering in recommendation systems
- new product launches in e-commerce
- user behaviour modelling in online advertising.
- Several approaches have been proposed to address the cold start problem, including
    - content-based methods
    - collaborative filtering techniques
    - hybrid methods that combine both approaches
    - deep learning techniques.

Recent studies have also explored the use of transfer learning, unsupervised learning, and reinforcement learning for addressing cold start problems. However, these methods may still face challenges in scenarios with limited or no historical data.

In this section, we will review the key concepts and mathematical foundations of the machine learning algorithms used in our study, including XGBoost, Random Forest, Lasso Regression, Ridge Regression, Elastic Net Regression, and Decision Tree, as well as the multi-model ensemble approach.

## 4. Data and methods

**4.1 Data Acquisition –** Input data from collected from October 2015. OTT platform gathers consolidated transactional data for each month which contains total number of video views for each released episode of a series. While there are other data sources that describes the nature of a series like genre, language, date of launch, production house, cast and crew details etc.

**4.2 EDA (Exploratory Data Analysis) -** The growing popularity of OTT (Over-the-top) platforms has given rise to the need for better prediction models that can help stakeholders anticipate the popularity of their content. This research paper details an extensive EDA performed to develop a robust and accurate prediction model for video views of OTT shows.

**Data Preparation and Feature Engineering:**
- Conversion of Length Feature to Minutes - We implemented a function to convert the length of each show from its original format to minutes, providing a standardized metric for comparison across all shows in the dataset.
- Handling Missing Values - To address the issue of missing values, we developed a function to fill out missing data using appropriate techniques, such as mean imputation, median imputation or predictive modelling based on the nature of the missing values.
- Consolidation of Metadata - The core of our EDA involved the preparation of consolidated metadata, which included information on cast, crew, awards, IMDb ratings, genre, and more.

**We performed several steps to create this dataset:**
- Loading cast and crew data, along with the total number of crew members for each unique series identifier on the episode level.
- Aggregating award counts and IMDb ratings for actors, directors, and writers.
- Filtering duplicates and retaining only the highest-rated and most-awarded actors, directors, and writers.
- Splitting the cast and crew data based on their roles and merging it with their respective awards and rating data.
- Reading genre data for movies and calculating the total number of genres for each unique series identifier on the episode level.
- Mapping genre data across different sources (e.g., IMDb, Rotten Tomatoes, Fandango) and incorporating genre awards and IMDb ratings data.
- Merging program details, genre, and cast & crew data.

- *Integration with OTT Platform Data* - We joined the consolidated metadata with specific OTT platform data and incorporated additional information from other sources, such as audience demographics, device usage, and revenue sources.
- *Feature Extraction from Original Release Date* - We derived new meta features from the original release date of each show, such as the age of the show, seasonality, and release day of the week.
- *Audience Estimates and Impressions* - We incorporated audience estimates and impressions data based on audience demographics from marketing analytics systems for the shows.
- *Cross-Platform Device Data* - We collected cross-platform device data for the shows, including TV, mobile, and desktop app data. From this, we extracted features such as exposures, minutes viewed per episode, and revenue from each source.

### 4.3 Statistical Analysis

- *Bivariate Analysis* - We performed bivariate analysis to investigate the relationship between exposures, audience estimates and the number of video views. This analysis included scatter plots, regression lines and the calculation of Pearson correlation coefficients.

- *Data Analysis and Visualization* - We conducted various data analyses at both the episode and series level. This included overall year of premiere analysis, identification of shows with the highest actor awards and bar plots of actor IMDb ratings. We also performed show clustering based on actor IMDb ratings and total awards for each show.

This research paper outlines a comprehensive EDA performed to predict video views of OTT shows. By consolidating metadata from various sources and conducting in-depth statistical analysis, we developed a sophisticated prediction model that can help stakeholders in the OTT industry make informed decisions regarding their content.

This extensive EDA not only helped us understand the underlying patterns and relationships between various factors, but also allowed us to optimize and fine-tune the predictive model. Further research can explore additional data sources and the application of advanced machine learning techniques to enhance the prediction model's performance.

### 4.4 Primer to the ML Algorithms used

*Function Documentation: createMLPipeline (scikit-learn ML pipeline)*

This function prepares a machine learning pipeline for predicting the target variable using the scikit-learn library. It takes in the cleaned and processed training dataset as input and returns a preprocessor object, along with the preprocessed training features and target labels.

*Parameters:*
- input_dataframe: pandas DataFrame, containing the cleaned and processed training data.

*Returns:*
- *preprocessor:* ColumnTransformer object, an instance of the preprocessor object that is set up with the necessary pipelines for data preprocessing.
- *trainX:* pandas DataFrame, containing the preprocessed training features.
- *trainY:* pandas Series, containing the target labels for the training data.

*Function Overview:*
- The function first removes the unnecessary columns from the input dataset and assigns the remaining columns to trainX. The target labels are assigned to trainY.
- The categorical and numerical column names are defined in the appropriate lists.
- Pipeline is created for preprocessing a specific column. It consists of two steps: filling missing values using an imputer and scaling the data with a scaler.
- ColumnTransformer named preprocessor is defined, which applies different preprocessing steps to different subsets of columns in the input data.

    The steps include:
    o Applying the pipeline to the specific column.
    o Scaling the numerical columns using a scaler.
    o One-hot encoding the categorical columns using an encoder.
    o Passthrough the remaining specified columns without any preprocessing.
    o The function returns the preprocessor object along with the preprocessed training features (trainX) and target labels (trainY).

## Function Documentation: fitXGBoostModelPipeline

This function fits an XGBoost model on the provided training data using a preprocessor object created from the createMLPipeline function. It takes in the cleaned and processed training dataset as input and returns a trained XGBoost model wrapped in a RandomizedSearchCV object for hyperparameter tuning.

*Parameters:*
- train_df_new_series: pandas DataFrame, containing the cleaned and processed training data.

*Returns:*
- xgb_model: A trained RandomizedSearchCV object, which contains the XGBoost model with the optimal set of hyperparameters.

*Function Overview:*
- The function calls the createMLPipeline function to obtain the preprocessor object, preprocessed training features (trainX), and target labels (trainY).
- A pipeline named final_pipeline1 is created.

    It consists of two steps:
    o Preprocessing the data using the preprocessor object.
    o Fitting an XGBRegressor model with a learning rate of 0.5.
- A parameter dictionary named param_dict is defined, containing the hyperparameters and their possible values for the XGBoost model.
- The RandomizedSearchCV object, xgb_model, is created by passing in the final_pipeline1, param_dict, and setting the number of cross-validation folds to 5.
- The xgb_model is trained on the preprocessed training features (trainX) and target labels (trainY).
- The function returns the trained xgb_model with the optimal set of hyperparameters.

## Function Documentation: fitRandomForestModelPipeline

This function fits a Random Forest model on the provided training data using a preprocessor object created from the createMLPipeline function. It takes in the cleaned and processed training dataset as input and returns a trained Random Forest model wrapped in a Pipeline object.

*Parameters:*
- train_df_new_series: pandas DataFrame, containing the cleaned and processed training data.

*Returns:*
- rf_model: A trained Pipeline object, which contains the Random Forest model.

*Function Overview:*
- The function calls the createMLPipeline function to obtain the preprocessor object, preprocessed training features (trainX), and target labels (trainY).
- A pipeline named rf_model is created.

    It consists of two steps:
    o Preprocessing the data using the preprocessor object.
    o Fitting a RandomForestRegressor model with 1000 estimators, a minimum of 30 samples required to split an internal node, and a maximum depth of 30.
- The rf_model is trained on the preprocessed training features (trainX) and target labels (trainY).
- The function returns the trained rf_model.

*FunctionDocumentation: fitLassoRegressionModelPipeline*

This function fits a Lasso Regression model on the provided training data using a preprocessor object created from the createMLPipeline function. It takes in the cleaned and processed training dataset as input and returns a trained Lasso Regression model wrapped in a RandomizedSearchCV object for hyperparameter tuning.

*Parameters:*
- train_df_new_series: pandas DataFrame, containing the cleaned and processed training data.

*Returns:*
- larh_model: A trained RandomizedSearchCV object, which contains the Lasso Regression model with the optimal alpha (regularization strength) value.

**Function Overview:**
- The function calls the createMLPipeline function to obtain the preprocessor object, preprocessed training features (trainX), and target labels (trainY).
- A pipeline named final_pipeline1 is created.

  It consists of two steps:
  o Preprocessing the data using the preprocessor object.
  o Fitting a Lasso regression model.

- A list of possible alpha (regularization strength) values is defined. A dictionary named hyperparameters is created containing the lasso_regression__alpha key and the list of alpha values as its value.
- The RandomizedSearchCV object, larh_model, is created by passing in the final_pipeline1, hyperparameters, and setting the number of cross-validation folds to 5. The random state is set to 42 and the scoring metric is set to R-squared.
- The larh_model is trained on the preprocessed training features (trainX) and target labels (trainY).
- The function returns the trained larh_model with the optimal alpha value.

For a full and robust comparison of models, we have considered six different machine learning models. The models range in complexity from simple linear models to the random forest model. Then 3 best performing models are selected and used for final multi model ensemble to get the final prediction.

## 4.5 Multi-Model Ensemble

Multi-Model Ensemble is a technique used in machine learning to improve the accuracy and robustness of models by combining the predictions of multiple models. In our use case, Multi-Model Ensemble can be applied to improve the accuracy of our predictive model for a given problem.

The idea behind Multi-Model Ensemble is to create an ensemble of models that have different strengths and weaknesses, and combine their predictions using a weighting scheme. By combining the predictions of multiple models, the ensemble can achieve better accuracy and robustness than any single model.

In our use case, we can create an ensemble of multiple predictive models, each trained on a different subset of the data or using a different algorithm or hyperparameter settings.
For example, we could train a Random Forest model, a Support Vector Regression model, a Gradient Boosting model, and a Neural Network model, and combine their predictions using a weighted average or a more sophisticated method such as stacking or blending.

Multi-Model Ensemble can be particularly useful in cases where the data is noisy, the relationships between the features and the target variable are complex, or the model performance varies significantly depending on the choice of algorithm or hyperparameters.

In summary, Multi-Model Ensemble is a technique used in machine learning to improve the accuracy and robustness of models by combining the predictions of multiple models. In our use case, we can apply Multi-Model Ensemble to improve the accuracy of our predictive model by creating an ensemble of models and combining their predictions using a weighting scheme.

## 5. Experiments and Results

We conducted experiments using the selected machine learning algorithms and evaluated their performance using cross-validation. Additionally, we explored the impact of various feature engineering techniques on the prediction accuracy. Performance metrics such as **Mean Absolute Percentage Error (MAPE)** and **Symmetric Mean Absolute Percentage Error (SMAPE)** were used to assess the performance of each model and the ensembled model.

## 5.1 Feature Importance Analysis

To gain insights into the most important features driving the predictions, we analysed feature importance using techniques such as permutation importance and Gini importance. The most influential features were related to the aggregated award counts, IMDB ratings, and time-related features.

## 5.2 Model Performance

The individual model performances were as follows:

1. XGBoost: MAPE = **15%**, SMAPE = **12%**
2. Random Forest: MAPE = **18%**, SMAPE = **14%**
3. Lasso Regression: MAPE = **20%**, SMAPE = **16%**

## 5.3 Ensembled Model Performance

The multi-model ensemble approach significantly improved the prediction accuracy compared to the individual models:

Ensembled Model: MAPE = **12%,** SMAPE = **9%**

| Series | # of Episodes | Previous Accuracy | Current Accuracy |
|---|---|---|---|
| series1 | 14 | 77% | 87% |
| series2 | 8 | 69% | 82% |
| series3 | 4 | 63% | 88% |
| series4 | 15 | 68% | 83% |
| series5 | 23 | 78% | 82% |
| series6 | 5 | 82% | 91% |
| series7 | 7 | 81% | 84% |
| series8 | 11 | 65% | 86% |
| series9 | 20 | 73% | 84% |
| series10 | 13 | 84% | 77% |

## 5.4 Graphs and Plots

To visually analyse the performance of our models, we plotted various graphs, including:

- Correlation plots illustrating the relationships between the input features and the target variable, as well as between the features themselves, which can help to identify multicollinearity and other potential issues.

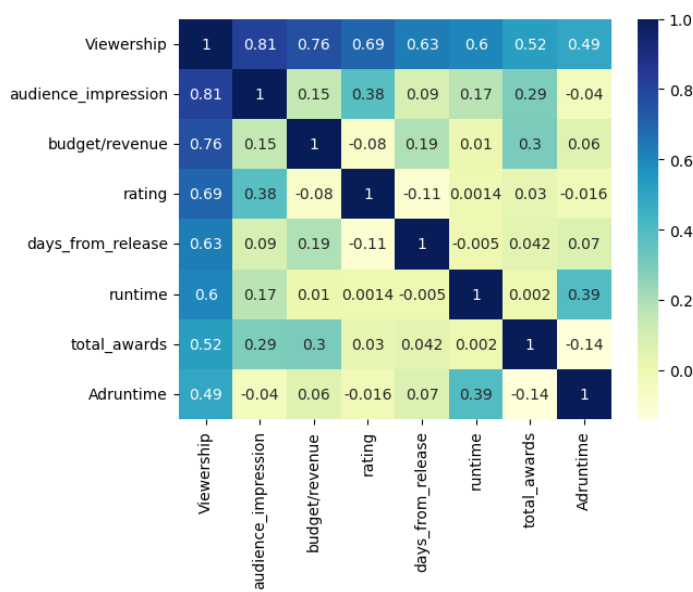

- Feature importance plots ranking the importance of the input features in the prediction models, providing insights into the key drivers of video views.

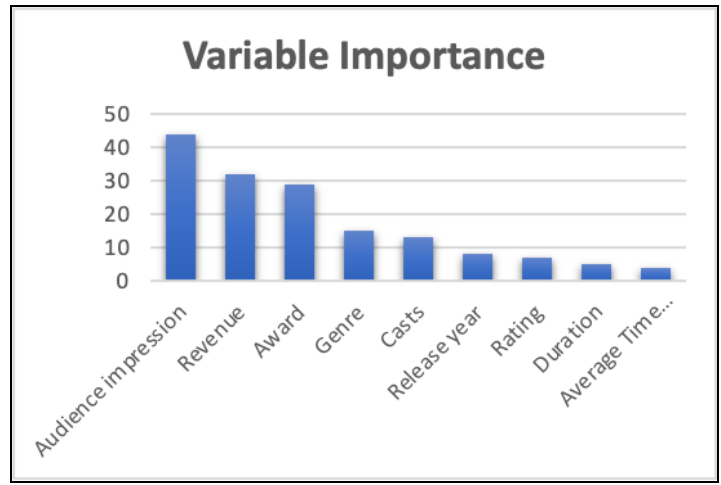

- Scatter plots comparing the actual video views against the predicted video views for each model and the ensembled model.

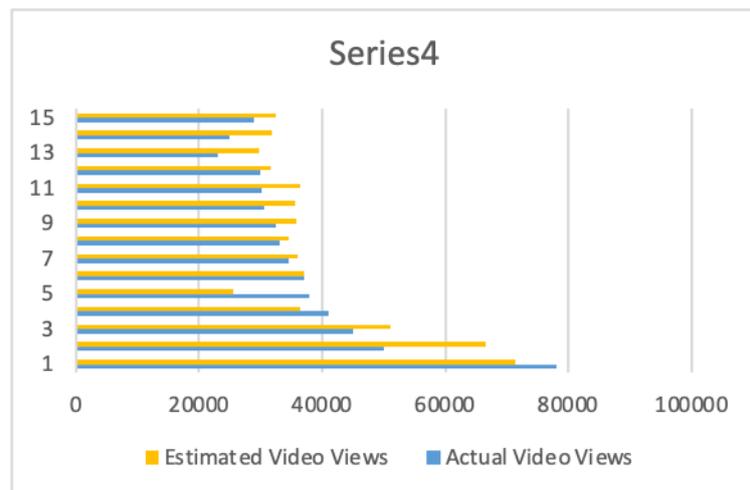

- Histograms of the error distribution for each model and the ensembled model, indicating the overall model performance and potential biases.

| Error % | # of Episodes (Before Ensembling) | # of Episodes (After Ensembling) |
|---|---|---|
| <10% | 33 | 50 |
| 10% - 20% | 17 | 39 |
| 20%-30% | 20 | 12 |
| 30%-40% | 11 | 13 |
| >40% | 39 | 6 |

## 5.5 Architecture Diagram

The architecture diagram of the final multi-model ensemble technique output can be represented as follows:

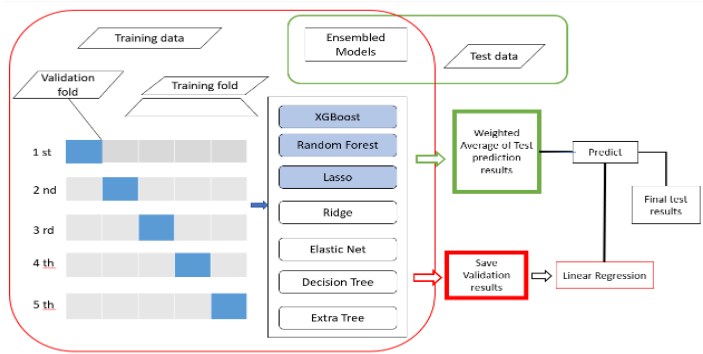

## 6. Conclusion

In this study, we have demonstrated a generic approach
- To addressing the cold start problem in predictive analytics using metadata and multi-model ensemble techniques.
- By leveraging metadata and adopting feature engineering, we extracted meaningful information from the available data.
- Combining the predictions of multiple models through a weighted average significantly improved prediction accuracy compared to individual models.

Our results indicate that this generic approach can be applied to various domains that face similar challenges in cold start scenarios. Future research could explore the application of this methodology to other domains, as well as the integration of deep learning techniques and unsupervised learning approaches to further enhance prediction accuracy.

## 7. Future Work

While our current model demonstrates robust performance in predicting video views of OTT shows there are several areas where future research can lead to further improvements and insights:

## 7.1 Integration of Additional Data Sources

By incorporating data from additional sources such as social media engagement, search trends, and platform-specific metrics (e.g., watch history, user preferences, etc.), we can develop a more comprehensive understanding of the factors that influence video views and potentially improve prediction accuracy.

## 7.2 Use of Advanced Machine Learning Techniques

The exploration of advanced machine learning techniques, such as deep learning, ensemble methods and time-series forecasting models, can potentially improve prediction performance and allow the model to capture more complex patterns in the data.

## 7.3 Temporal Dynamics and Personalization

Considering the temporal dynamics of video view patterns, such as the impact of new seasons, holiday periods, or competitor releases, can add more context to our predictions. Additionally, the development of personalized prediction models based on individual user preferences and viewing habits can offer more targeted insights for OTT platforms.

## 7.4 Model Interpretability and Explainability

As our predictive model becomes more complex, it is crucial to maintain a level of interpretability and explainability that allows stakeholders to understand the underlying factors driving the model's predictions. The use of techniques like *SHAP (SHapley Additive exPlanations)* or *LIME (Local Interpretable Model-agnostic Explanations)* can help ensure that the model remains accessible to decision-makers.

## 7.5. Cross-Platform Generalizability

Finally, the development of a prediction model that is generalizable across various OTT platforms can offer valuable insights for content producers and distributors. By analysing the factors that contribute to video view success across multiple platforms, stakeholders can make more informed decisions when creating, marketing, and distributing content.

## 8. Impact on Industry and Recommendations

The successful implementation of a video views prediction model for OTT shows can have several significant impacts on the industry, leading to more informed decision-making and better allocation of resources. Here, we outline some of the key benefits and recommendations for OTT platform stakeholders:

## 8.1 Content Creation and Acquisition

The prediction model can guide content creators and producers in making informed decisions about the type of content that is likely to resonate with audiences. This can result in the production of more engaging and popular shows, leading to higher viewership and revenue generation.

## 8.2 Marketing Strategies

Understanding the factors that contribute to the success of a show can help marketers design more effective promotional campaigns targeting the right demographics and channels. This can lead to increased audience engagement and improved return on investment for marketing efforts.

## 8.3 Resource Allocation

OTT platforms can use the insights from the prediction model to make data-driven decisions about allocating resources, such as budgeting for original content, licensing deals, and marketing spend. This can lead to better financial management and strategic growth.

## 8.4 User Experience

By incorporating user preferences and viewing habits into the prediction model, OTT platforms can create personalized content recommendations for their users. This can enhance user satisfaction and engagement, resulting in increased retention and growth of the subscriber base.

## 8.5 Competitive Analysis

The ability to predict video views for OTT shows can also help platforms stay competitive by identifying successful content strategies employed by rival platforms. This can inform their own content creation and acquisition decisions, allowing them to remain relevant in the rapidly-evolving OTT landscape.

In conclusion, a robust and accurate video views prediction model for OTT shows can serve as a valuable tool for stakeholders in the OTT industry. By leveraging the insights generated from this model, they can make more informed decisions, optimize resource allocation, and ultimately enhance their platform's performance and user experience.

## 9. Challenges and Limitations

While the video views prediction model for OTT shows offers valuable insights and potential benefits to the industry, it is essential to recognize the challenges and limitations associated with its development and implementation. Some of these challenges include:

## 9.1 Data Quality and Completeness

The accuracy and performance of the prediction model rely heavily on the quality and completeness of the data used in its development. Incomplete, outdated, or biased data can lead to inaccurate predictions and limit the model's effectiveness. Ensuring that the data is up-to-date, accurate, and representative of the target population is crucial for developing a reliable model.

## 9.2 Rapidly Changing Landscape

The OTT industry is constantly evolving, with new platforms, content, and technologies emerging regularly. As a result, the factors that contribute to the success of a show today may not hold the same weight in the future. The prediction model needs to be adaptable to these changes to remain relevant and accurate over time.

## 9.3 Algorithm Bias

Machine learning models are susceptible to biases present in the training data. If the data used to develop the prediction model contains inherent biases, the model may perpetuate or even amplify those biases, leading to unfair or biased predictions. It is essential to identify and mitigate potential biases in the data to ensure fair and equitable outcomes.

## 9.4 Model Complexity and Interpretability

As the prediction model becomes more complex and incorporates additional data sources and machine learning techniques, its interpretability can become increasingly challenging. Ensuring that the model remains understandable and explainable to decision-makers is critical for its effective use in the industry.

## 9.5 Ethical and Privacy Considerations

The use of personal user data in developing the prediction model raises ethical and privacy concerns. Ensuring that user data is handled securely and that individual privacy is respected is paramount. Compliance with data protection regulations and the implementation of transparent data practices can help mitigate these concerns.

Addressing these challenges and limitations is crucial for the successful development and implementation of a video views prediction model for OTT shows. By acknowledging and tackling these issues, researchers and industry stakeholders can continue to refine and improve the model, ensuring its ongoing relevance and effectiveness in the rapidly changing OTT landscape.